\RequirePackage{fix-cm}
\documentclass[smallextended]{svjour3}       
\smartqed  

\usepackage{graphicx}
\usepackage{amsmath}
\usepackage{amssymb}
\usepackage{algorithmic}
\usepackage{algorithm}
\usepackage{subfigure}
\usepackage{array}
\usepackage{tabularx}
\usepackage{multirow}

\newcommand{\bfx}{{\textbf{x}}}
\newcommand{\bfv}{{\textbf{v}}}

\newcommand{\bfw}{{\textbf{w}}}

\newcommand{\bfxi}{{\boldsymbol{\xi}}}

\newcommand{\bfalpha}{{\boldsymbol{\alpha}}}

\newcommand{\bfepsilon}{{\boldsymbol{\epsilon}}}
\newcommand{\bfdelta}{{\boldsymbol{\delta}}}
\newcommand{\bfgamma}{{\boldsymbol{\gamma}}}

\begin{document}

\title{Supervised learning of sparse context reconstruction coefficients for data representation and classification}

\author{Xuejie Liu \and Jingbin Wang \and Ming Yin \and Benjamin Edwards \and  Peijuan Xu}

\institute{
X. Liu, P. Xu\at
College of Computer Science and Technology, Jilin University, Changchun, 130012, China\\
\email{xuejie.liu@hotmail.com}
\and
J. Wang\at
National Time Service Center, Chinese Academy of Sciences, Xi' an 710600 , China\\
Graduate University of Chinese Academy of Sciences, Beijing 100039, China\\
\email{jingbinwang1@outlook.com}
\and
M. Yin, B. Edwards\at
Department of Computer Science, Sam Houston State University, Huntsville, TX 77341, USA\\
\email{ming.yin@hotmail.com}, \email{benjamin.edwards1@hotmail.com}
}

\date{Received: date / Accepted: date}

\maketitle

\begin{abstract}
Context of data points, which is usually defined as the other data points in a data set, has been found to paly important roles in data representation and classification. In this paper, we study the problem of using context of a data point for its classification problem. Our work is inspired by the observation that actually only very few data points are critical in the context of a data point for its representation and classification. We propose to represent a data point as the sparse linear combination of its context, and learn the sparse context in a supervised way to increase its discriminative ability. To this end, we proposed a novel formulation for context learning, by modeling the learning of context parameter and classifier in a unified objective, and optimizing it with an alternative strategy in an iterative algorithm. Experiments on three benchmark data set show its advantage over  state-of-the-art context-based data representation and classification methods.
\end{abstract}

\section{Introduction}
\label{sec:intro}

Pattern classification is a major problem in machine learning research \cite{Wang20151,Xu20121205,Guo20121893,He2013793,Wang201585,Tian20141007,wang2015representing}. This problem is defined as a problem of predicting a binary class label of a given data point. There are many examples of this problem in real-world applications. For example, in computer vision area, given an image of face, we may want to predict whose face it is \cite{Feng20151407,Xu2015307,Wang201575,Lu20151371,Zhao2015677,Li20152736,Wang2012963}. In natural language processing applications, given a text, we also want to predict which topic it is about \cite{Sheydaei20153,Nayak2015497,Agerri201536,Karad20157721,Li2015427,Garanina2015140}. Moreover, in applications of wireless sensor network, it is important to detect if one node is normal or at fault.
To solve this problem, we usually first represent the data point as a feature vector, and then learn a classifier function to predict the class label from its feature vector. The two most important topics of pattern classification are data representation and classifier learning. Most data representation and classification methods are based on single data point. When one data point is considered for representation and classification, all other data points are ignored. For example, in the most popular data representation method, feature selection scheme, when we have a feature vector a one data point, we simply reduce the abandoned features, and re-organize the remaining feature to a new feature vector to obtain the representation of the data point \cite{Dess20154632,Jin2015172}. In this procedure, no other data points are considered beside the data point to represent. Another example is the most classification method, support vector machine (SVM). When we have a test, a linear function is applied to its feature vector to predict its class label \cite{Kim201519,Kang20152786}. In this procedure, no other data points are considered. However, the other data points other than the data point under consideration may play important roles in its representation and classification. These data points are called ``context" of the considered data point. A data point may have different true nature in different context. Thus it is necessary to explore the contexts of data points when they are represented and/or classified. To this end, some methods have been proposed to use the context of a data point for its representation and classification. In this paper, we investigate the problem of learning effective representation of a data point from its context guided by its class label, and proposed a novel supervised context learning method using sparse regularization and linear classifier learning formulation.

\subsection{Related works}

This paper is to explore the context information for data representation and classification, thus we give some brief review of existing context-based data representation and classification methods.

\begin{itemize}
\item The most popular context-based data classification is $k$ nearest neighbor classification (KNN).  Given a test data point and a training set, we first search the training set to find the $k$ nearest neighbors of the test data point to present its context, and then we determine its class label by a majority vote of the the labels of the context \cite{Ahn201563,aldea2014classifications}. All the data points of the context contribute equally to the final classification result, and no representation procedure is needed.

\item Wright et al. \cite{wright2009robust} proposed sparse representation based classification (SRBC),
to use the data points of one class as a context of a test data point, and reconstruct it by its context. The reconstruction coefficients are imposed to be sparse. Moreover, the class with the minimum reconstruction error is assigned to the test data point. This method does not require to learn an explicate classifier to predict the class label. Thus it cannot take advantages of the classifier learning technologies.

\item Melacci and Belkin \cite{melacci2011laplacian} proposed Laplacian support vector machine (LSVM), to use the $k$ nearest neighbors of a training data point to present its context, and learn a linear classifier to respect the context. Specifically, the classification result of a training data point is imposed to be similar to its contextual data points. However, after the classifier is trained, and used to classify a test data point, the context of the test data point is ignored.

\item Gao et al. \cite{gao2010local} proposed Laplacian sparse coding (LSC) to represent the context of a data point by using its $k$ nearest neighbors, and represent the data points with regard to the contexts. Each data point is reconstructed as a linear combination of the codewords of a dictionary, and the combination coefficients are imposed to be sparse. Moreover, the combination coefficients of a data point are impose to be similar to these of its contextual data points. This method is unsupervised simply a data representation method, and the class label information is ignored.
\end{itemize}

\subsection{Contributions}

We propose a novel method to explore the context of a data point, and use it to represent it. Moreover, a linear classifier function is learned to predict its class label from its representation based on its context. We use its $k$ nearest neighbors as its context, and try to reconstruct it by the data points in its context. The reconstruction errors are imposed to be spares, and we measure the sparsity by a $\ell_1$ norm regularization, similar to sparse coding \cite{Wang20141630,Liu20151452,Wang2013199,Staglian20152415,Wang20149}. Moreover, the reconstruction result is used as the new representation of this data point. We apply a linear function to predict its class label. To learn the reconstruction coefficient vectors of the data points and the classifier parameter vector, we build a unified objective function. In this function, the reconstruction error are measured by a squared $\ell_2$ norm distance, and the classification error is measured by the hinge loss. Moreover, the $\ell_1$ norm regularization is applied to the reconstruction coefficient vectors to encourage their sparsity, and the squared $\ell_2$ norm regularization is applied to the classifier parameter vector to reduce the complexity of the classifier. By optimizing the objective function with regard to both the reconstruction coefficient vectors and the classifier parameter vector, the context based representation and classier are learned simultaneously. In this way, the context and the classifier can regularize the learning of each other. To minimize the proposed objective function, we use the Lagrange multiplier and an alternate optimization method, and develop an iterative algorithm based on the optimization results. The contributions of this paper are of two folds:

\begin{enumerate}
\item We propose a novel context representation formulation. A data point is represented by its sparse reconstruction of its context. The motivation of this contribution is that for each data point, only a few data points in its context is of the same class as itself. However, it is critical to find which data points plays the most important roles in its context for the classification of the data point itself. To find the critical contextual data points, we proposed to learn the classifier together with the sparse context. The classifier can be used to regularize the learning of the reconstruction coefficient vector, and thus find the critical data points in the context. We mode this problem as a minimization problem. In this problem, the context reconstruction error, reconstruction sparsity, classification error, and classifier complexity are minimized simultaneously.

\item We also problem a novel iterative algorithm to solve this minimization problem. We first reformulate it as its Lagrange formula, and the use an alterative optimization method to solve it. In each iteration, we first fix the classifier parameter vector to update the reconstruction vectors, and then fix the reconstruction vectors to  update the classifier parameter vector.
\end{enumerate}

\subsection{Paper organization}

This paper is organized as follows. In section \ref{sec:method}, we introduce the proposed method. In section \ref{sec:experiment}, we evaluate the proposed method experimentally. In section \ref{sec:conclusion}, this paper is concluded with future works.

\section{Proposed method}
\label{sec:method}

In this section, we introduce the proposed classification method which explores
the context information.
The learning problem is firstly formulated by modeling an objective function,
and then it is optimized in an iterative algorithm.

\subsection{Problem formulation}

We consider a binary classification problem, and a training set of $n$ data points are given as
$\{\bfx_i\}_{i=1}^n$, where $\bfx_i\in \mathbb{R}^d$ is a $d$-dimensional feature vector of the $i$-th data point.
The binary class labels of the training points are given as $\{y_i\}_{i=1}^n$ and $y_i\in \{+1,-1\}$ is
the class label of the $i$-th point.
To learn from the context of the $i$-th data point, we find its $k$ nearest neighbors and denote
them as $\{\bfx_{ij}\}_{j=1}^k$, where $\bfx_{ij}$ is the $j$-th nearest neighbor of the $i$-th point.
They are further organized as a $d\times k$ matrix
$X_i=[\bfx_{i1},\cdots,\bfx_{ik} ] \in R^{d\times {k}}$, where the $j$-th column is $\bfx_{ij}$.
The $k$ nearest neighbors of the $i$-th point is used to represent its context information.
We represent $\bfx_i$ by linearly reconstructing it from its contextual points as

\begin{equation}
\begin{aligned}
\bfx_i \approx \widehat{\bfx}_i
= \sum_{j=1}^k \bfx_{ij} v_{ij}
= X_i \bfv_i
\end{aligned}
\end{equation}
where $\widehat{\bfx}_i$ is its reconstruction, and $v_{ij}$ is the reconstruction coefficient
of the $j$-th nearest neighbor.
$\bfv_i=[v_{i1},\cdots,v_{ik}]^\top \in \mathbb{R}^k$ is the reconstruction coefficient vector of the $i$-th data point. The reconstruction coefficient vectors of all the training points are organized in reconstruction coefficient matrix
$V=[\bfv_1,\cdots,\bfv_n] \in \mathbb{R}^{k\times n}$, with its $i$-th column as $\bfv_i$.
The key idea of this method is an assumption the for both the reconstruction and classification
of $\bfx_i$, only a few of its nearest neighbors play important role, while the remaining neighbors
could be discarded, resulting a sparse context.
To encourage the sparsity of the context, we impose a $\ell_1$ norm penalty to the
contextual reconstruction coefficient vector $\bfv_i$.
Moreover, to learn the contextual reconstruction coefficient vectors, we also propose to minimized the reconstruction error measured by a squared $\ell_2$ norm penalty between $\bfx_i$ and $X_i \bfv_i$, and the following optimization problem is obtained,

\begin{equation}
\label{equ:reconstruction}
\begin{aligned}
\min_{V}
~
&\left \{
\beta \sum_{i=1}^n  \|\bfx_i - X_i \bfv_i\|_2^2 +  \gamma \sum_{i=1}^n  \|\bfv_i\|_1 \right \},
\end{aligned}
\end{equation}
where $\beta$ and $\gamma$ are trade-off parameters.

To classify $\bfx_i$, instead of applying a classifier to $\bfx_i$ itself, we apply a linear classifier to its
contextual reconstruction $\widehat{\bfx}_i$.
The classifier is defined as

\begin{equation}
\begin{aligned}
f(\widehat{\bfx}_i) = \bfw^\top  \widehat{\bfx}_i
=\bfw^\top X_i \bfv_i
\end{aligned}
\end{equation}
where $\bfw\in \mathbb{R}^d$ is the classifier parameter vector.
To learn the classifier, we consider the hinge-loss function and
the squared $\ell_2$ norm regularization simultaneously.
The following optimization problem is obtained with regard to the classifier learning,

\begin{equation}
\label{equ:classification}
\begin{aligned}
\min_{\bfw,V,\bfxi}
~
&\left \{ \frac{1}{2}\| \bfw \|^2_2 +
\alpha \sum_{i=1}^n  \xi_i \right \}\\
s.t.~
&
1- y_i\left (\bfw^\top X_i \bfv \right ) \leq \xi_i, \xi_i \geq 0, i=1,\cdots, n,
\end{aligned}
\end{equation}
where $\frac{1}{2}\| \bfw \|^2_2 $ is the the squared $\ell_2$ norm regularization term to reduce the complexity
of the classifier, $\xi_i$ is the slack variable for the hinge loss of the $i$-th training point,
$\bfxi=[\xi_1,\cdots,\xi_n]^\top$
and $\alpha$ is a tradeoff parameter.

The overall optimization problem is obtained by combining the problems in both (\ref{equ:reconstruction})
and (\ref{equ:classification}) as

\begin{equation}
\label{equ:objective}
\begin{aligned}
\min_{\bfw,V,\bfxi}
~
&\left \{ \frac{1}{2}\| \bfw \|^2_2 +
\alpha \sum_{i=1}^n  \xi_i + \beta \sum_{i=1}^n  \|\bfx_i - X_i \bfv_i\|_2^2 +  \gamma \sum_{i=1}^n  \|\bfv_i\|_1
\right \} \\
s.t.~
&
1- y_i\left (\bfw^\top X_i \bfv \right ) \leq \xi_i, \xi_i \geq 0, i=1,\cdots, n.
\end{aligned}
\end{equation}
From the above problem, we can see that by encouraging the sparsity of $\bfv_i$, we learn a sparse context for
both the reconstruction and classification of $\bfx_i$.

\subsection{Optimization}

To optimize the constrained problem in (\ref{equ:objective}), we write the Lagrange function of this problem as

\begin{equation}
\begin{aligned}
\mathcal{L}(\bfw,V,\bfxi,\bfgamma,\bfdelta)=
&
\frac{1}{2}\| \bfw \|^2_2 +
\alpha \sum_{i=1}^n  \xi_i + \beta \sum_{i=1}^n  \|\bfx_i - X_i \bfv_i\|_2^2 +  \gamma \sum_{i=1}^n  \|\bfv_i\|_1\\
&
+\sum_{i=1}^n \delta_i \left(1- y_i\left (\bfw^\top X_i \bfv_i \right ) - \xi_i \right)
-\sum_{i=1}^n \epsilon_i \xi_i,
\end{aligned}
\end{equation}
where $\delta_i $ is the Lagrange multiplier for the constrain of $1- y_i\left (\bfw^\top X_i \bfv \right ) \leq \xi_i$,
and $\epsilon_i$ is the Lagrange multiplier for the constrain of $\xi_i\geq 0$.
According to the dual theory of optimization, the following dual optimization problem is obtained,

\begin{equation}
\label{equ:dual}
\begin{aligned}
\max_{\bfdelta,\bfepsilon}&\min_{\bfw,V,\bfxi}
~
\mathcal{L}(\bfw,\bfv,\bfxi,\bfdelta,\bfepsilon)\\
s.t.~
&
\bfdelta\geq 0,\bfepsilon\geq 0,
\end{aligned}
\end{equation}
where $\bfdelta=[\delta_1,\cdots,\delta_n]^\top$, and $\bfepsilon=[\epsilon_1,\cdots,\epsilon_n]^\top$.
By setting the partial derivative of $\mathcal{L}$ with regard to $\bfw$ to zero, we have

\begin{equation}
\label{equ:w}
\begin{aligned}
&\frac{\partial \mathcal{L}}{\partial \bfw}=0
\Rightarrow
\bfw = \sum_{i=1}^n \delta_i  y_i  X_i \bfv_i.
\end{aligned}
\end{equation}
By setting the partial derivative of $\mathcal{L}$ with regard to $\xi_i$ to zero, we have

\begin{equation}
\label{equ:xi}
\begin{aligned}
\frac{\partial \mathcal{L}}{\partial \xi_i}=0
&\Rightarrow
\alpha-\delta_i-\epsilon_i=0\\
&\Rightarrow
\alpha-\delta_i=\epsilon_i\\
\epsilon_i\geq 0
&\Rightarrow
\alpha \geq \delta_i.
\end{aligned}
\end{equation}
Substituting (\ref{equ:w}) and (\ref{equ:xi}) to (\ref{equ:dual}),
we eliminate $\bfw$ and $\bfdelta$

\begin{equation}
\label{equ:objective1}
\begin{aligned}
\max_{\bfdelta}&\min_{V}
~
\left \{
-\frac{1}{2} \sum_{i,j=1}^n \delta_i  \delta_j y_i y_j \bfv_i ^\top X_i^\top   X_j \bfv_j
+ \beta \sum_{i=1}^n  \|\bfx_i - X_i \bfv_i\|_2^2
\right. \\
&\left .+  \gamma \sum_{i=1}^n  \|\bfv_i\|_1
+\sum_{i=1}^n \delta_i
\right \}
\\
s.t.~
&
\bfalpha \geq \bfdelta \geq 0.
\end{aligned}
\end{equation}
where $\bfalpha=[\alpha,\cdots,\alpha]^\top$ is a $n$ dimensional vector of all $\alpha$ elements.
It is difficult to solve this dual problem
with a close form solution. We try to solve it with the
alternate optimization strategy.
In each iteration of an iterative algorithm, we fix $\bfdelta$ first to solve $V$,
and then fix $V$ to solve $\bfdelta$.

\subsubsection{Solving $V$ while fixing $\bfdelta$}

When $\bfdelta$ is fixed and only $V$ is considered, the problem in (\ref{equ:objective1}) is reduced to

\begin{equation}
\label{equ:V}
\begin{aligned}
\min_{V}
~
& \left \{
-\frac{1}{2} \sum_{i,j=1}^n \delta_i  \delta_j y_i y_j \bfv_i ^\top X_i^\top   X_j \bfv_j
+ \beta \sum_{i=1}^n  \|\bfx_i - X_i \bfv_i\|_2^2 +  \gamma \sum_{i=1}^n  \|\bfv_i\|_1
\right \}.
\end{aligned}
\end{equation}
Instead of solving $V$ at one time, we solve $\bfv_i|_{i=1}^n$ one by one.
When the contextual reconstruction vector of the $i$-th point $\bfv_i$ is considered, we
fix that of all other points $\bfv_i|_{j\neq i}$.
(\ref{equ:V}) is further reduced to

\begin{equation}
\label{equ:v}
\begin{aligned}
\min_{\bfv_i}
~
&\left \{
-\frac{1}{2} \sum_{i,j=1}^n \delta_i  \delta_j y_i y_j \bfv_i ^\top X_i^\top   X_j \bfv_j
+ \beta   \|\bfx_i - X_i \bfv_i\|_2^2 +  \gamma   \|\bfv_i\|_1 \right \}.
\end{aligned}
\end{equation}
This problem could be solved efficiently by the modified feature-sign search algorithm
proposed by Gao et al. \cite{gao2013laplacian}.

\subsubsection{Solving $\bfdelta$ while fixing $V$}

When $V$ is fixed and only $\bfdelta$ is considered, the problem in (\ref{equ:objective1})
is reduced to

\begin{equation}
\label{equ:delta}
\begin{aligned}
\max_{\bfdelta}
~
&
\left \{
-\frac{1}{2} \sum_{i,j=1}^n  \delta_i  \delta_j y_i y_j \bfv_i ^\top X_i^\top   X_j \bfv_j
+\sum_{i=1}^n \delta_i \right \}\\
s.t.~
&
\bfalpha \geq \bfdelta \geq 0.
\end{aligned}
\end{equation}
This problem is a typical constrained quadratic programming (QP) problem, and it can be solved efficiently
by the active set algorithm.

\subsection{Iterative algorithm}

The iterative algorithm to learn both the classifier parameter $\bfw$ and the
contextual reconstruction coefficient vectors in $V$ is given in Algorithm 1.
As we can see from the algorithm, the iterations are repeated $T$ times and then the
updated $V$ and $\bfdelta$ are outputs. Please note that the variables of this algorithm are initialized randomly.

\begin{description}
\item[Algorithm 1: Iterative Learning algorithm.]~

\begin{description}
\item[Input] Training point set $\{\bfx_i\}_{i=1}^n$ and label set $\{y_i\}_{i=1}^n$;
\item[Input] Nearest neighbor size parameter $k$;
\item[Input] Tradeoff parameters $\alpha$, $\beta$ and $\gamma$;
\item[Input] Maximum iteration number $T$.

\item[Initialization] Find nearest neighbors $\{\bfx_{ij}\}_{j=1}^k$ for each data point $\bfx_i, i=1,\cdots, n$.

\item[Initialization] Initialize $\bfdelta^0$ randomly;

\item[For $t=1,\cdots T$] ~

\begin{enumerate}
\item Fix $\bfdelta^{t-1}$ and update the contextual reconstruction coefficient vectors $\bfv_i^t|_{i=1}^n$ one by one by solving the problem in (\ref{equ:v});
\item Fix $\bfv_i^t|_{i=1}^n$  and update the classifier parameter vector $\bfw^t$ by solving
$\bfdelta^{t}$ as in (\ref{equ:delta}).
\end{enumerate}

\item[Endfor]

\item[Output] classifier parameter vector $\bfw = \sum_i^n \delta_i^T y_i X_i \bfv_i^T$ .

\end{description}

\end{description}

\subsection{Classifying a test point}

When a new test point $\bfx \in \mathbb{R}^d$ comes, to represent its context, we also find its $k$ nearest neighbors
from the training set  and put them in a $d \times k $ matrix $X$. Given a classifier parameter vector $\bfw$, and a candidate class label $y\in \{+1,-1\}$, we seek its class conditional context reconstruction coefficient vector, by solving the following minimization problem,

\begin{equation}
\begin{aligned}
\bfv^y=
\underset{\bfv}{\arg\min}
~
&
\left \{ -y\bfw^\top (X \bfv) + \beta \|\bfx - X \bfv\|_2^2 +  \gamma \|\bfv\|_1 \right \}.
\end{aligned}
\end{equation}
This problem can also be solved by the modified feature-sign search algorithm
proposed by Gao et al. \cite{gao2013laplacian}. The final class label $y^*$ of the test data point is obtained as the candidate label minimizing the following objective,

\begin{equation}
\begin{aligned}
y^*=
\min_{y\in \{+1,-1\}}
\left \{
-y\bfw^\top (X \bfv^y)
\right \}.
\end{aligned}
\end{equation}

\section{Experiments}
\label{sec:experiment}

In this section, we evaluate the proposed supervised sparse context learning (SSCL) algorithm on several benchmark data sets.

\subsection{Data sets}

In the experiments, we used three date sets, which are introduced as follows:

\begin{itemize}
\item \textbf{MANET loss data set}:
The packet losses of the receiver in mobile Ad hoc networks (MANET) can be classified into three types, which are wireless random errors caused losses, the route change losses induced by node mobility and network congestion. It is very important to recognize which class a packet loss belongs in research and application of mobile Ad hoc networks. The first data set used in our experiments is a MANET loss data set. To construct this data set, we simulate a MANET scenario by using a network simulator NS-2 \cite{Guerreiro2014213,Pouria20148633}. We put 30 nodes in a $400m\times 800m$ area, select a TFRC flow as the observation stream, and a TCP flow as the background traffic between two randomly selected nodes. The random error rate is confided from 1\% to 10\%.
We collect 381 data points for the congestion loss, 458 for the route change loss, and 516 data points for the wireless error loss. Thus in the data set, there are 1355 data points in total. To extract the feature vector each data point, we calculate 12 features from each data point as in \cite{Deng2009}, and concatenate them to form a vector.

\item \textbf{Twitter data set}: The second data set is a Twitter data set. The target of this data set is to predict the gender of the twitter user, male or female, given one of his/her Twitter massage. To construct this data set, we downloaded Twitter massages of 50 male users and 50 female users of 100 days. We collected 53,971 twitter massages in total, and among them there are 28,012 messages sent by male users, and 25,959 messages sent by female users. To extract features from each Twitter message, we extract Term features, linguistic features, and medium diversity features as gender-specific features as in  \cite{Huang2014488}.

\item \textbf{Arrhythmia data set}: The third data set is publicly available at http://arc\\
    hive.ics.uci.edu/ml/datasets/Arrhythmia. In this data set, there are 452 data points, and they belongs to 16 different classes. Each data point has a feature vector of 279 features.

\end{itemize}

\subsection{Experiment setup}

To conduct the experiments, we used the 10-fold cross validation. A entire data set is split into 10 folds, and each of them was used as a test set in turn. The remaining 9 folds are combined and used as a training set. The learning algorithm was applied to the training set to learn the classifier parameter. The algorithm is adjusted by using a 9-fold cross validation on the training set. The learned classifier was then applied to the test set to predict the class labels of the testing data points. The prediction performance is evaluated by the prediction accuracy, which is defined as,

\begin{equation}
\begin{aligned}
Prediction~ accuracy=
\frac{Number~of~correctly~predicted~testing~data~points}{Total~number~of~testing~data~points}.
\end{aligned}
\end{equation}

\subsection{Results}

In the experiments, we first compare the proposed context-based data representation and classification algorithm, SSCL, to several context-based data representation and/or classification methods. Then we study the sensitivity of the proposed algorithm to its parameters experimentally. Finally, we study the convergency of the proposed iterative algorithm.

\subsubsection{Comparison to context-based representation and classification methods}

\begin{figure}[!htb]
  \centering
  \subfigure[MANET loss data set]{
  \includegraphics[width=0.48\textwidth]{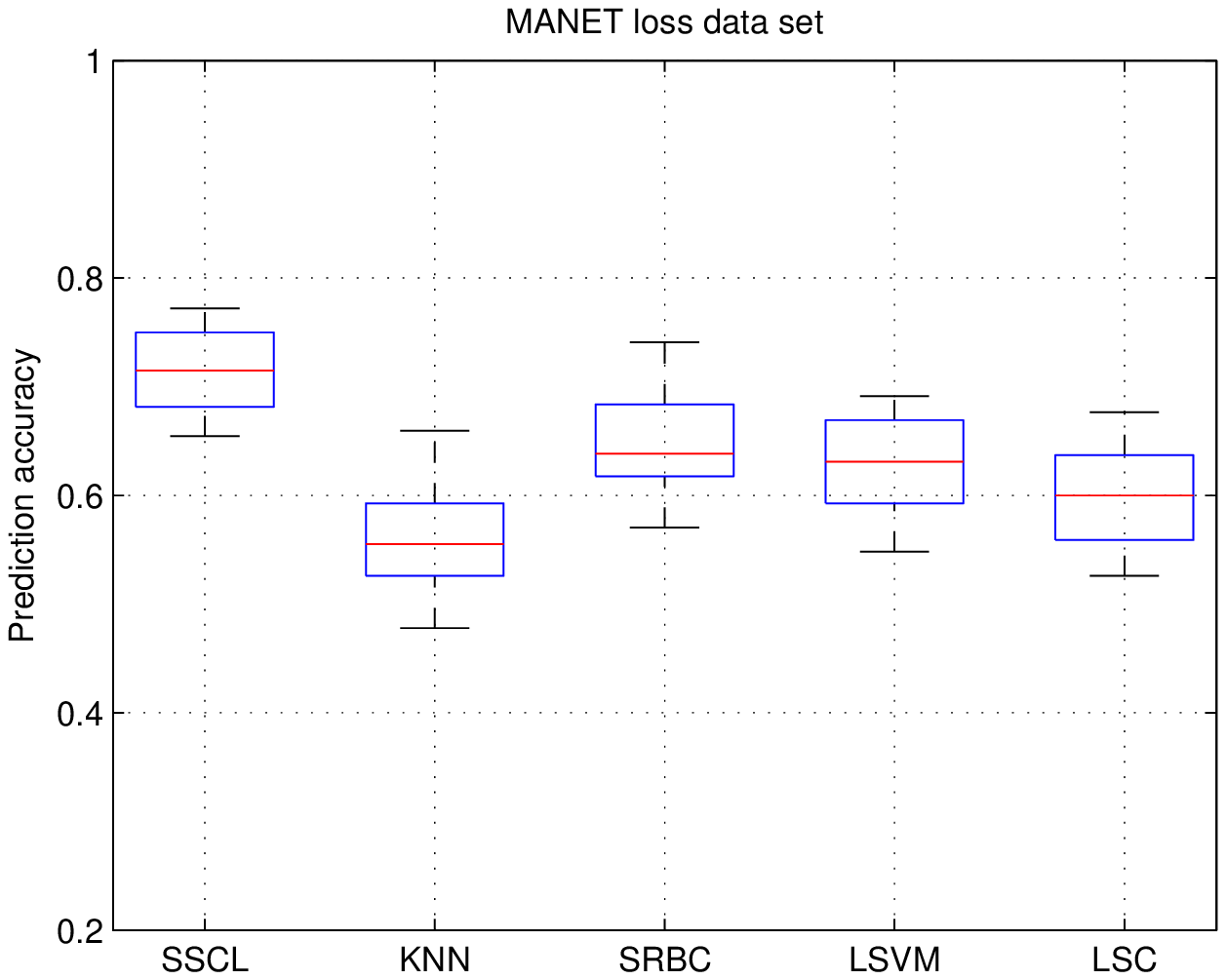}}
  \subfigure[Twitter data set]{
  \includegraphics[width=0.48\textwidth]{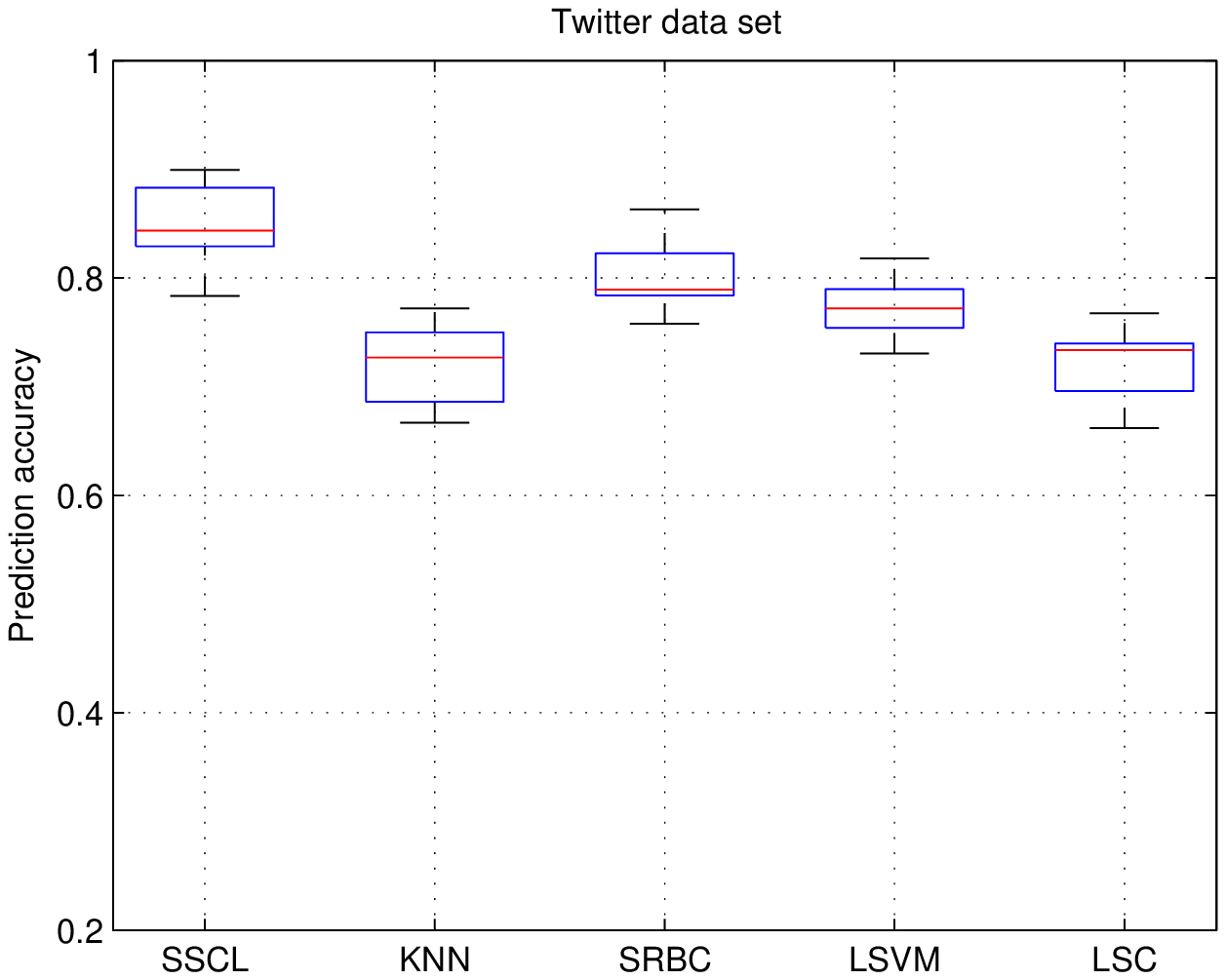}}
  \subfigure[Arrhythmia data set]{
  \includegraphics[width=0.48\textwidth]{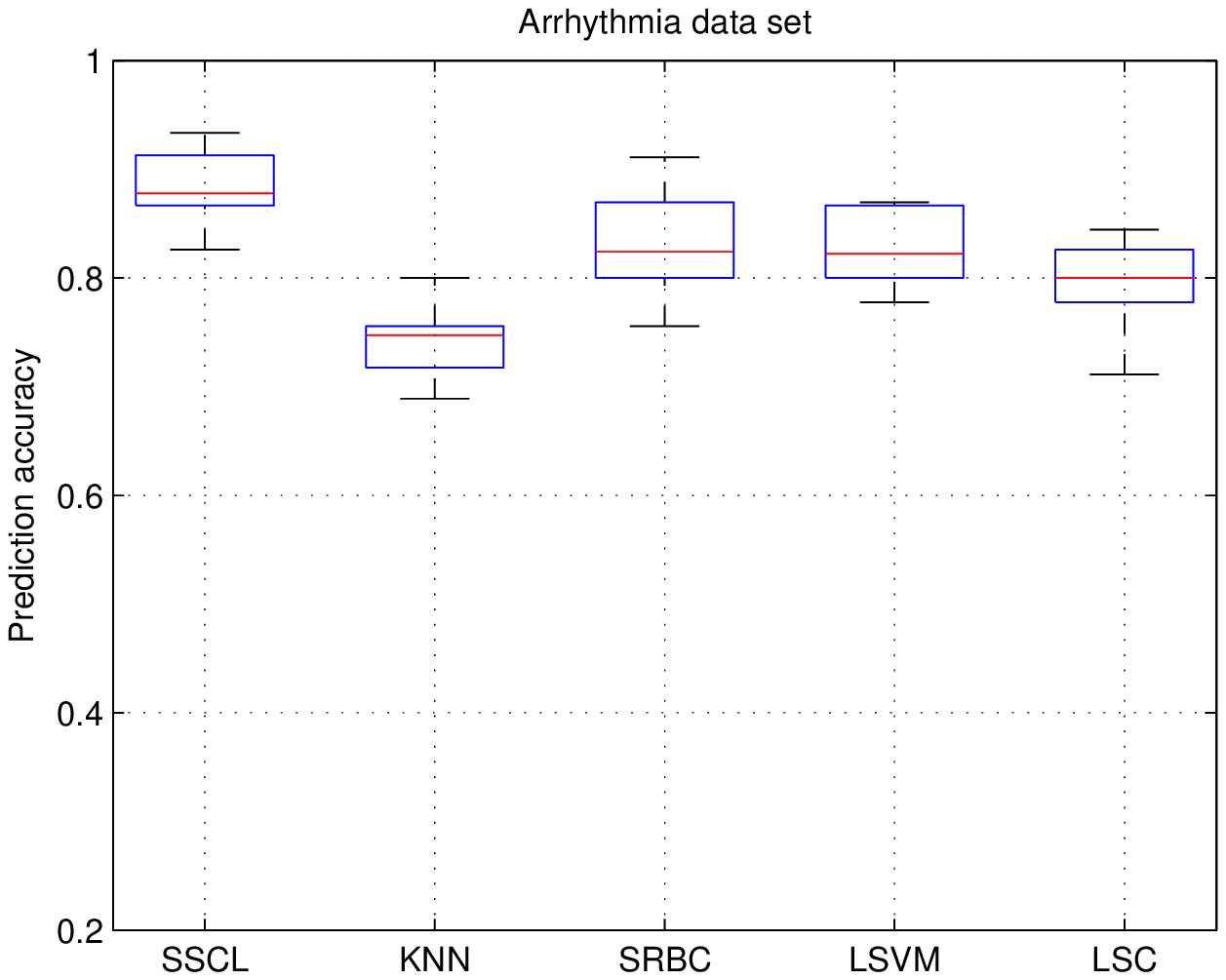}}
  \caption{Boxplots of prediction accuracy of different context-based algorithms.}
  \label{fig:compare}
\end{figure}

Since the proposed algorithm is a context-based classification and sparse representation method, we compared the proposed algorithm to three popular context-based classifiers, and one context-based sparse representation method.
The three context-based classifiers are traditional KNN, Wright et al.'s SRBC \cite{wright2009robust}, and Melacci and Belkin's LSVM \cite{melacci2011laplacian}. The context-based sparse representation method is Gao et al.'s LSC \cite{gao2010local}. The boxplots of the 10-fold cross validation of the compared algorithms are given in figure \ref{fig:compare}. From the figures, we can see that the proposed method SSCL outperforms all the other methods on all three data sets. Among median values of the boxplots of prediction accuracies over three data sets, SSCL are always the highest one. In most cases, the 25-th percentiles of SSCL is even higher than the median values of other algorithms. The second best method is SRBC, which also uses sparse context to represent the data point. However, compared to SSCL, it doesn't learn any explicit classifier for the classification problem. Thus it cannot take advantage of the classifier design tricks. This is the mean reason that SRBC inferior to SSCL. KNN also uses context to classify a data point without using a explicit classifier. However, unlike SRBC whose context is class-conditional, KNN uses a general context and treats all contextual data points equally, and obtains the worst classification results. This is a strong evidence that learning a supervised sparse context is critical for classification problem. LSVM also uses context information to regularize the learning of classifier. However, once the classifier is learned, the context is ignored in the classification procedure, thus its performance is inferior to SSCL. LSC is an unsupervised learning algorithm, and it is not surprising that its performance is not good.

\subsubsection{Sensitivity to parameters}

\begin{figure}[!htb]
  \centering
\subfigure[$\alpha$]{
\label{fig:alpha}
\includegraphics[width=0.48\textwidth]{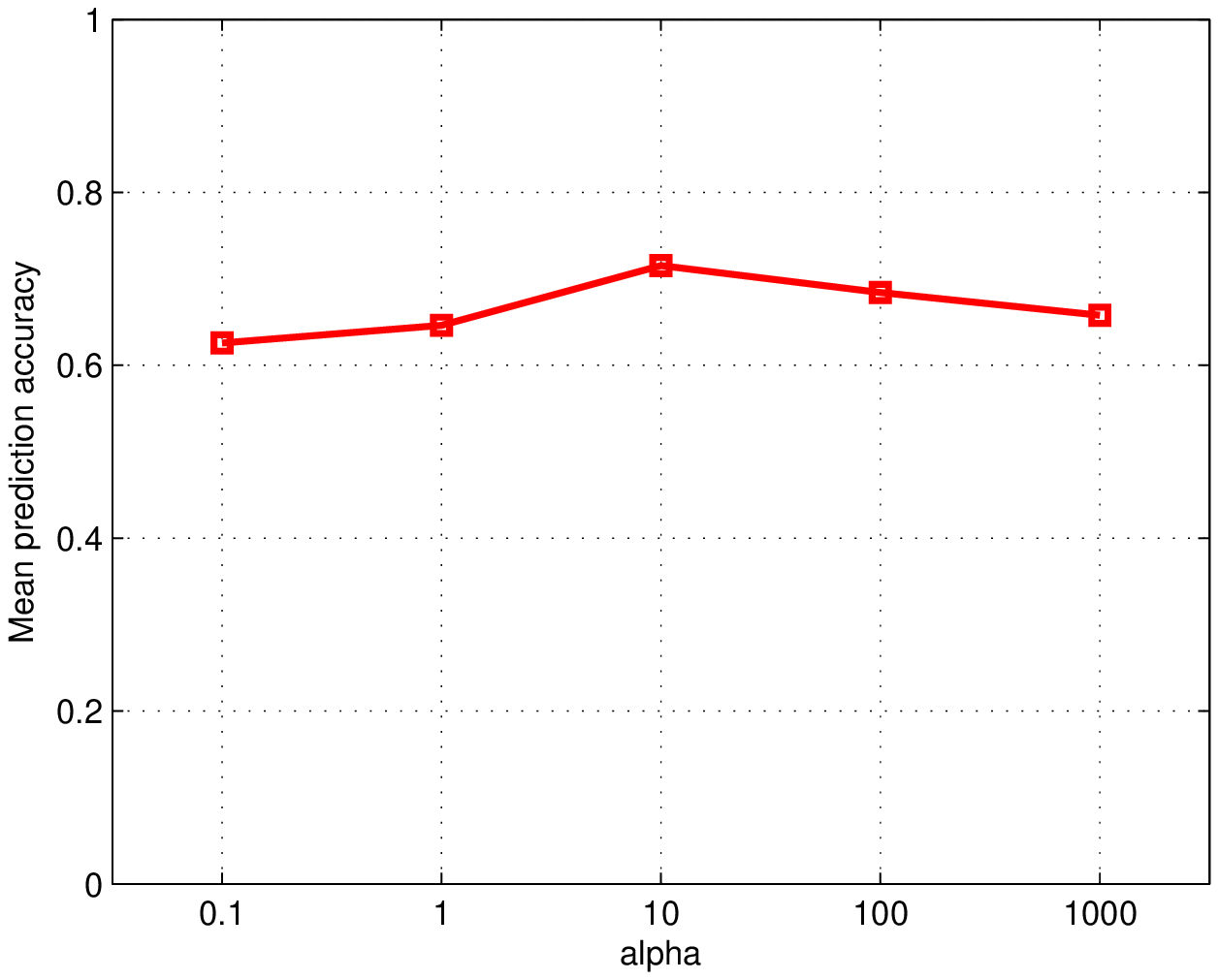}}
\subfigure[$\beta$]{
\label{fig:beta}
\includegraphics[width=0.48\textwidth]{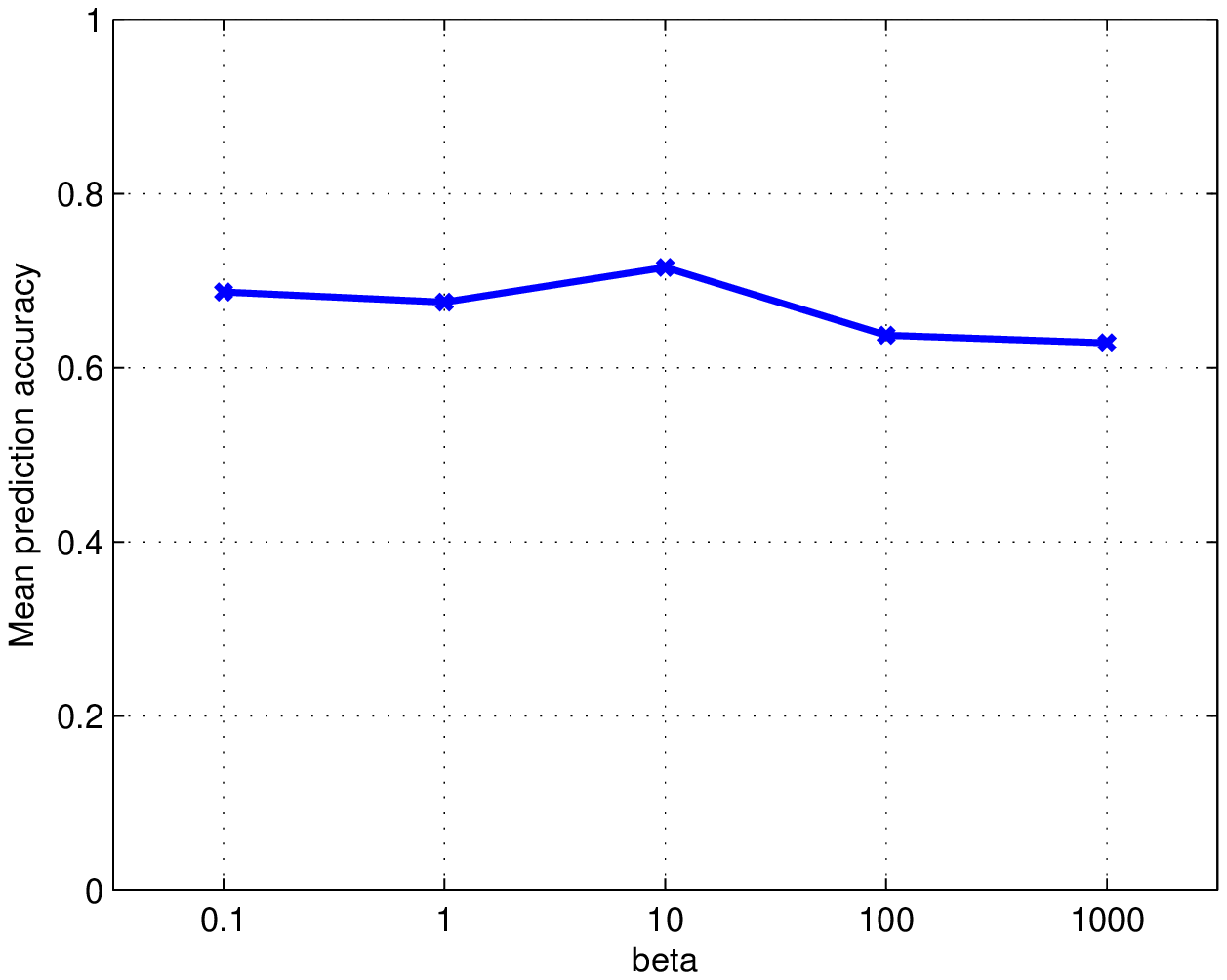}}
\subfigure[$\gamma$]{
\label{fig:gamma}
\includegraphics[width=0.48\textwidth]{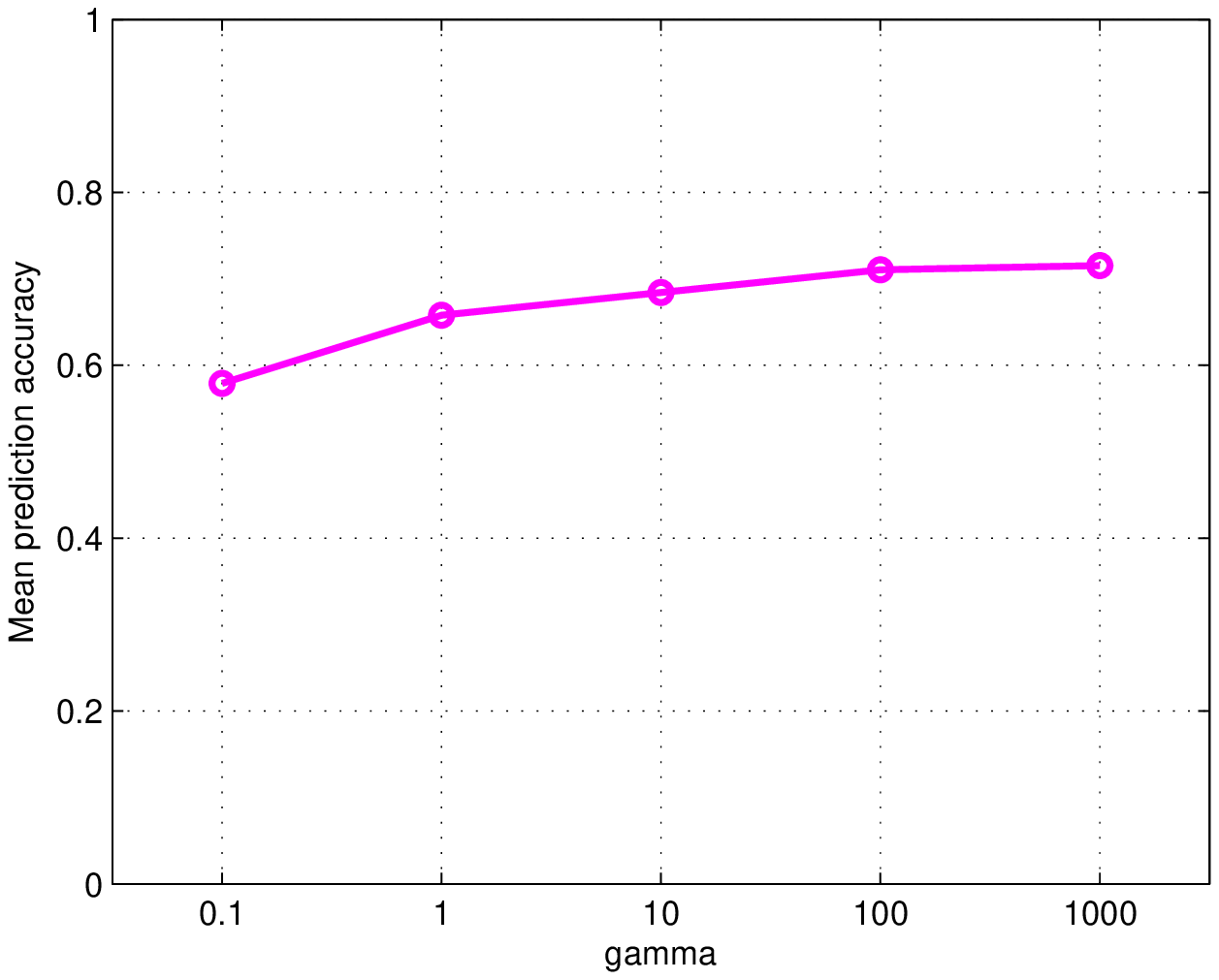}}
\subfigure[$k$]{
\label{fig:k}
\includegraphics[width=0.48\textwidth]{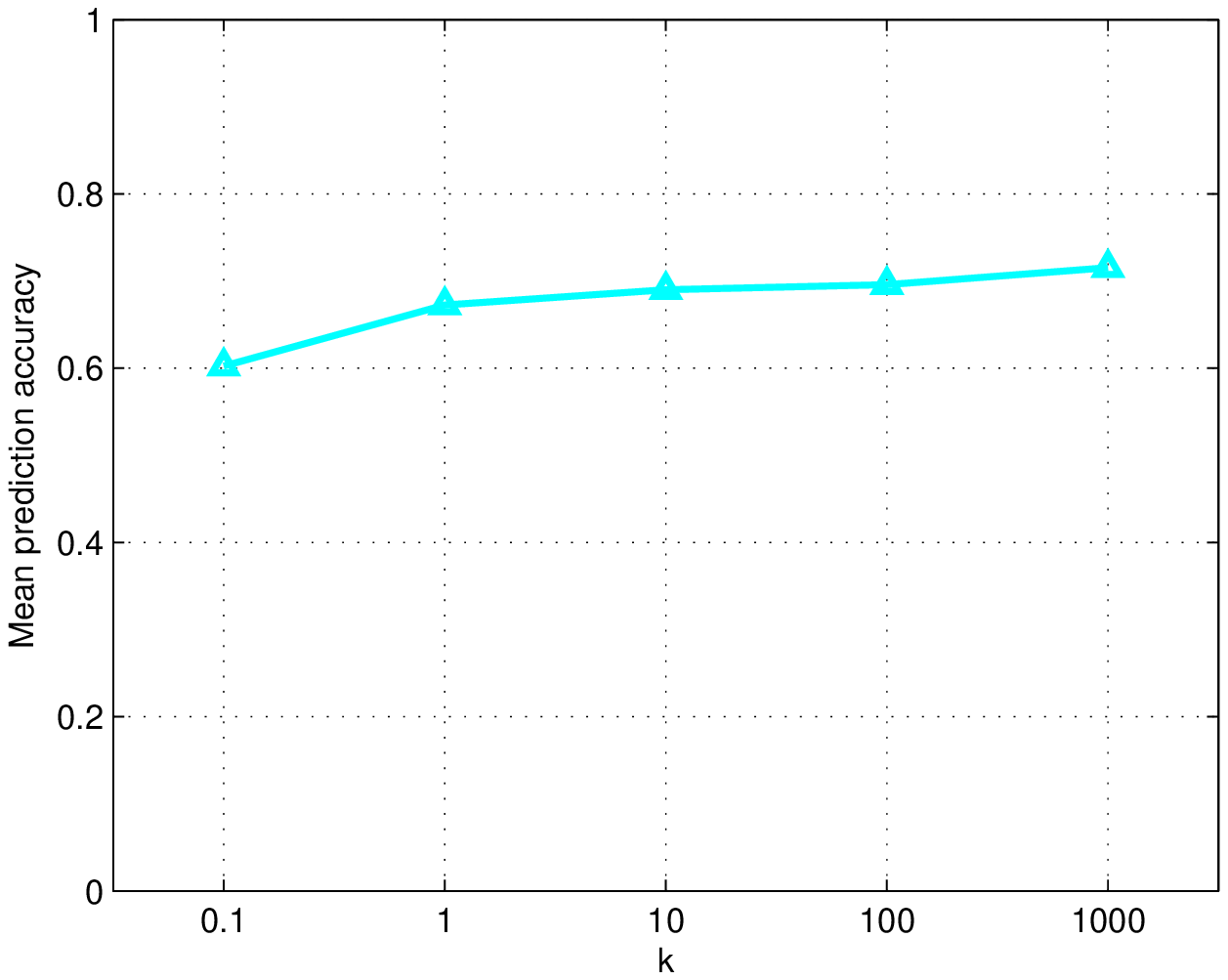}}
\\
\caption{Parameter sensitivity curves.}
  \label{fig:para}
\end{figure}

In the proposed formulation, there are three tradeoff parameters, $\alpha$, $\beta$, and $\gamma$. Moreover, we have one more parameter, which is the size of the neighborhood, $k$. It is interesting to investigate how these parameters effects the performance of the proposed algorithm. We plot the curve of mean prediction accuracies against different values of parameters, and show them in figure \ref{fig:para}. From figure \ref{fig:alpha} and \ref{fig:beta}, we can see the accuracy is stable to the parameter $\alpha$ and $\beta$. More specifically, in figure \ref{fig:alpha}, it seems that the performances are a little better with a median value of $\alpha$. $\alpha$ is the weight of the hinge loss function, and when it makes sense the classifier has a better performance with a median value, since a too large values leads to over-fitting, while a too small value leads to training error over the training set.
It is also interesting to note that $\beta$ also achieves the best performance with a median value, 10. $\beta$ is the weight of the reconstruction error term. A small weight of this term makes the representation of a data point irrelevant to itself, while a large weight does not grantee its discriminative ability. From figure \ref{fig:gamma} and \ref{fig:k}, we can see a larger $\gamma$ or $k$ leads to better classification performances. $\gamma$ is the weight of the sparsity term, a larger $\gamma$ achieves a higher prediction accuracy means prediction result benefits from a sparsity representation. This is because that in the context of a data point, only a few data points plays important roles. Sparsity of the context forces the model to select those important contextual data points. $k$ is the size of the context, and a larger $k$ provides more candidate contextual data points, and helps the model to find the critical contextual data points.

\subsubsection{Algorithm convergency}

\begin{figure}[!htb]
  \centering
  \includegraphics[width=0.8\textwidth]{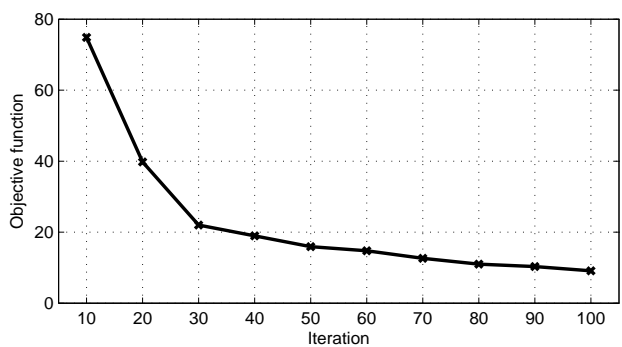}\\
  \caption{Convergency curve of the proposed SSCL algorithm.}
  \label{fig:convergency}
\end{figure}

We are also interested in the convergency of the proposed iterative algorithm SSCL. We plot the objective function of the formulation in (\ref{equ:objective}) in different iterations, and show the convergency curve in figure \ref{fig:convergency}. From this figure, it is clear that the algorithm converge after the $50$-th iteration.

\subsubsection{Running time analysis}

\begin{figure}[!h]
  \centering
  \includegraphics[width=0.8\textwidth]{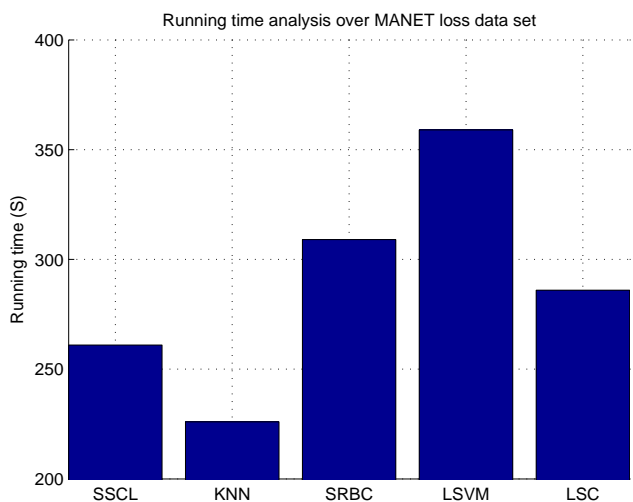}\\
  \caption{Running time of different algorithms over MANET loss data set.}\label{FigRunningtime150809}
\end{figure}

We also provide an analysis of the running time of the compared algorithms over the MANET loss data set. The running time of the algorithms is given in figure \ref{FigRunningtime150809}. The unit of the running time is second. From the figure, we can see that the least time consuming algorithm is KNN, however, its classification performance is poor.  Our algorithm, SSCL, is the second least time consuming algorithm. It takes no more than 250 seconds, while all other algorithms take more than that. Moreover, SSCL achieves the best classification results. It leads to the conclusion that the proposed algorithm can achieve the best classification performance with a reasonable running time.

\section{Conclusion and future works}
\label{sec:conclusion}

In this paper, we study the problem of using context to represent and classify data points. Our motivation is that although data points in context of a data point plays important roles in its classification, only a few of them is critical. Thus it is necessary to learn a sparse context. To this end, we propose to use a sparse linear combination of the data points in the context of a data point to represent itself. Moreover, to increase the discriminative ability of the new representation, we develop an supervised method to learn the sparse context by learning it and a classifier together in an unified optimization framework. Experiments on three benchmark data sets show its advantage over state-of-the-art context-based data representation and classification methods.

Although the proposed method works well for small data sets, it cannot scale up to large data set. The reason is that in each iteration, it solves a QP problem with regard to the number of data points in (\ref{equ:delta}). This procedure works with small number of data points, however, when it is large, it is too consuming to solve such a QP problem with so many variables. In the future, we will investigate to release this QP problem to a linear problem, by using the expectation-maximization (EM) framework to release the hinge loss to a linear function. Moreover, we also plan to extend the proposed algorithm to different applications, e.g., bioinformatics \cite{Wang2013dsf,Wang2012564,Wang2012dsfgdsf,Wang2014258}, computer vision \cite{wang2014effective,Wang20133249,Duan20158,Tom201580}, and information retrieval \cite{Wang2014635,Jasiewicz201562,Greving2015291,Wang2013}.

\section*{Acknowledgements}

This work was supported by the Fundamental Research Funds of Jilin University, China, (Grant No. 450060491509).


\begin{thebibliography}{10}
\providecommand{\url}[1]{{#1}}
\providecommand{\urlprefix}{URL }
\expandafter\ifx\csname urlstyle\endcsname\relax
  \providecommand{\doi}[1]{DOI~\discretionary{}{}{}#1}\else
  \providecommand{\doi}{DOI~\discretionary{}{}{}\begingroup
  \urlstyle{rm}\Url}\fi

\bibitem{Agerri201536}
Agerri, R., Artola, X., Beloki, Z., Rigau, G., Soroa, A.: Big data for natural
  language processing: A streaming approach.
\newblock Knowledge-Based Systems \textbf{79}, 36--42 (2015)

\bibitem{Ahn201563}
Ahn, W.H., Nah, S.P., Seo, B.S.: Automatic classification of digitally
  modulated signals based on k-nearest neighbor.
\newblock Lecture Notes in Electrical Engineering \textbf{329}, 63--69 (2015)

\bibitem{aldea2014classifications}
Aldea, R., Fira, M., Lazar, A.: Classifications of motor imagery tasks using
  k-nearest neighbors.
\newblock In: Neural Network Applications in Electrical Engineering (NEUREL),
  2014 12th Symposium on, pp. 115--120. IEEE (2014)

\bibitem{Deng2009}
Deng, Q., Cai, A.: Svm-based loss differentiation mechanism in mobile ad hoc
  networks.
\newblock In: 2009 Global Mobile Congress, GMC 2009 (2009).
\newblock \doi{10.1109/GMC.2009.5295834}

\bibitem{Dess20154632}
Dess{\'i}, N., Pes, B.: Similarity of feature selection methods: An empirical
  study across data intensive classification tasks.
\newblock Expert Systems with Applications \textbf{42}(10), 4632--4642 (2015)

\bibitem{Duan20158}
Duan, Y., Stien, L., Thorsen, A., Karlsen, {\o}., Sandlund, N., Li, D., Fu, Z.,
  Meier, S.: An automatic counting system for transparent pelagic fish eggs
  based on computer vision.
\newblock Aquacultural Engineering \textbf{67}, 8--13 (2015)

\bibitem{Feng20151407}
Feng, Q., Pan, T., Pan, J., Tang, L.: Improved mean representation based
  classification for face recognition.
\newblock Lecture Notes in Electrical Engineering \textbf{330}, 1407--1412
  (2015)

\bibitem{gao2013laplacian}
Gao, S., Tsang, I.H., Chia, L.T.: Laplacian sparse coding, hypergraph laplacian
  sparse coding, and applications.
\newblock IEEE Transactions on Pattern Analysis and Machine Intelligence
  \textbf{35}(1), 92--104 (2013)

\bibitem{gao2010local}
Gao, S., Tsang, I.W., Chia, L.T., Zhao, P.: Local features are not
  lonely--laplacian sparse coding for image classification.
\newblock In: Computer Vision and Pattern Recognition (CVPR), 2010 IEEE
  Conference on, pp. 3555--3561. IEEE (2010)

\bibitem{Garanina2015140}
Garanina, N., Sidorova, E.: Ontology population as algebraic information system
  processing based on multi-agent natural language text analysis algorithms.
\newblock Programming and Computer Software \textbf{41}(3), 140--148 (2015)

\bibitem{Greving2015291}
Greving, H., Sassenberg, K.: Counter-regulation online: Threat biases retrieval
  of information during internet search.
\newblock Computers in Human Behavior \textbf{50}, 291--298 (2015)

\bibitem{Guerreiro2014213}
Guerreiro, A., Souza, J., Rufino, J.: Improving ns-2 network simulator to
  evaluate ieee 802.15.4 wireless networks under error conditions.
\newblock pp. 213--220 (2014)

\bibitem{Guo20121893}
Guo, Z., Li, Q., You, J., Zhang, D., Liu, W.: Local directional derivative
  pattern for rotation invariant texture classification.
\newblock Neural Computing and Applications \textbf{21}(8), 1893--1904 (2012)

\bibitem{He2013793}
He, Y., Sang, N.: Multi-ring local binary patterns for rotation invariant
  texture classification.
\newblock Neural Computing and Applications \textbf{22}(3-4), 793--802 (2013)

\bibitem{Huang2014488}
Huang, F., Li, C., Lin, L.: Identifying gender of microblog users based on
  message mining.
\newblock Lecture Notes in Computer Science (including subseries Lecture Notes
  in Artificial Intelligence and Lecture Notes in Bioinformatics) \textbf{8485
  LNCS}, 488--493 (2014)

\bibitem{Jasiewicz201562}
Jasiewicz, J., Netzel, P., Stepinski, T.: Geopat: A toolbox for pattern-based
  information retrieval from large geospatial databases.
\newblock Computers and Geosciences \textbf{80}, 62--73 (2015)

\bibitem{Jin2015172}
Jin, C., Jin, S.W.: Automatic image annotation using feature selection based on
  improving quantum particle swarm optimization.
\newblock Signal Processing \textbf{109}, 172--181 (2015)

\bibitem{Kang20152786}
Kang, M., Kim, J., Kim, J.M., Tan, A., Kim, E., Choi, B.K.: Reliable fault
  diagnosis for low-speed bearings using individually trained support vector
  machines with kernel discriminative feature analysis.
\newblock IEEE Transactions on Power Electronics \textbf{30}(5), 2786--2797
  (2015)

\bibitem{Karad20157721}
Karad, A., Joshi, R.: Rule based chunk extraction from pdf documents using
  regular expressions and natural language processing.
\newblock International Journal of Applied Engineering Research \textbf{10}(3)
  (2015)

\bibitem{Kim201519}
Kim, S., Yu, Z., Kil, R., Lee, M.: Deep learning of support vector machines
  with class probability output networks.
\newblock Neural Networks \textbf{64}, 19--28 (2015)

\bibitem{Li2015427}
Li, H., Federico, M., He, X., Meng, H., Trancoso, I.: Introduction to the
  special section on continuous space and related methods in natural language
  processing.
\newblock IEEE Transactions on Audio, Speech and Language Processing
  \textbf{23}(3), 427--430 (2015)

\bibitem{Li20152736}
Li, Z., Gong, D., Li, X., Tao, D.: Learning compact feature descriptor and
  adaptive matching framework for face recognition.
\newblock IEEE Transactions on Image Processing \textbf{24}(9), 2736--2745
  (2015)

\bibitem{Liu20151452}
Liu, T., Wang, G., Wang, L., Chan, K.: Visual tracking via temporally smooth
  sparse coding.
\newblock IEEE Signal Processing Letters \textbf{22}(9), 1452--1456 (2015)

\bibitem{Lu20151371}
Lu, J., Liong, V., Wang, G., Moulin, P.: Joint feature learning for face
  recognition.
\newblock IEEE Transactions on Information Forensics and Security
  \textbf{10}(7), 1371--1383 (2015)

\bibitem{melacci2011laplacian}
Melacci, S., Belkin, M.: Laplacian support vector machines trained in the
  primal.
\newblock The Journal of Machine Learning Research \textbf{12}, 1149--1184
  (2011)

\bibitem{Nayak2015497}
Nayak, M., Nayak, A.: Odia running text recognition using moment-based feature
  extraction and mean distance classification technique.
\newblock Advances in Intelligent Systems and Computing \textbf{309
  AISC}(VOLUME 2), 497--506 (2015)

\bibitem{Pouria20148633}
Pouria, Z., Mathews, E., Havinga, P., Stojanovski, S., Sisinni, E., Ferrari,
  P.: Implementation of wirelesshart in the ns-2 simulator and validation of
  its correctness.
\newblock Sensors (Switzerland) \textbf{14}(5), 8633--8668 (2014)

\bibitem{Sheydaei20153}
Sheydaei, N., Saraee, M., Shahgholian, A.: A novel feature selection method for
  text classification using association rules and clustering.
\newblock Journal of Information Science \textbf{41}(1), 3--15 (2015)

\bibitem{Staglian20152415}
Staglian{\'o}, A., Noceti, N., Verri, A., Odone, F.: Online space-variant
  background modeling with sparse coding.
\newblock IEEE Transactions on Image Processing \textbf{24}(8), 2415--2428
  (2015)

\bibitem{Tian20141007}
Tian, Y., Zhang, Q., Liu, D.: v-nonparallel support vector machine for pattern
  classification.
\newblock Neural Computing and Applications \textbf{25}(5), 1007--1020 (2014)

\bibitem{Tom201580}
Tom{\'e}, A., Kuipers, M., Pinheiro, T., Nunes, M., Heitor, T.: Space-use
  analysis through computer vision.
\newblock Automation in Construction \textbf{57}, 80--97 (2015)

\bibitem{wang2014effective}
Wang, H., Wang, J.: An effective image representation method using kernel
  classification.
\newblock In: Tools with Artificial Intelligence (ICTAI), 2014 IEEE 26th
  International Conference on, pp. 853--858 (2014)

\bibitem{Wang2013}
Wang, J., Gao, X., Wang, Q., Li, Y.: Prodis-contshc: Learning protein
  dissimilarity measures and hierarchical context coherently for
  protein-protein comparison in protein database retrieval.
\newblock BMC Bioinformatics \textbf{13}(SUPPL.7) (2013)

\bibitem{Wang2012564}
Wang, J., Li, Y., Wang, Q., You, X., Man, J., Wang, C., Gao, X.: Proclusensem:
  Predicting membrane protein types by fusing different modes of pseudo amino
  acid composition.
\newblock Computers in Biology and Medicine \textbf{42}(5), 564--574 (2012)

\bibitem{wang2015representing}
Wang, J., Zhou, Y., Yin, M., Chen, S., Edwards, B.: Representing data by sparse
  combination of contextual data points for classification.
\newblock In: Advances in Neural Networks--ISNN 2015. Springer (2015)

\bibitem{Wang2012dsfgdsf}
Wang, J.J.Y., Bensmail, H., Gao, X.: Multiple graph regularized protein domain
  ranking.
\newblock BMC Bioinformatics \textbf{13}(1) (2012)

\bibitem{Wang20133249}
Wang, J.J.Y., Bensmail, H., Gao, X.: Joint learning and weighting of visual
  vocabulary for bag-of-feature based tissue classification.
\newblock Pattern Recognition \textbf{46}(12), 3249--3255 (2013)

\bibitem{Wang20149}
Wang, J.J.Y., Bensmail, H., Gao, X.: Feature selection and multi-kernel
  learning for sparse representation on a manifold.
\newblock Neural Networks \textbf{51}, 9--16 (2014)

\bibitem{Wang2013199}
Wang, J.J.Y., Bensmail, H., Yao, N., Gao, X.: Discriminative sparse coding on
  multi-manifolds.
\newblock Knowledge-Based Systems \textbf{54}, 199--206 (2013)

\bibitem{Wang20141630}
Wang, J.J.Y., Gao, X.: Semi-supervised sparse coding.
\newblock In: Proceedings of the International Joint Conference on Neural
  Networks, pp. 1630--1637 (2014)

\bibitem{Wang201575}
Wang, J.J.Y., Gao, X.: Max-min distance nonnegative matrix factorization.
\newblock Neural Networks \textbf{61}, 75--84 (2015)

\bibitem{Wang2014635}
Wang, J.J.Y., Sun, Y., Gao, X.: Sparse structure regularized ranking.
\newblock Multimedia Tools and Applications \textbf{74}(2), 635--654 (2014)

\bibitem{Wang2013dsf}
Wang, J.J.Y., Wang, X., Gao, X.: Non-negative matrix factorization by
  maximizing correntropy for cancer clustering.
\newblock BMC Bioinformatics \textbf{14} (2013)

\bibitem{Wang201585}
Wang, J.J.Y., Wang, Y., Jing, B.Y., Gao, X.: Regularized maximum correntropy
  machine.
\newblock Neurocomputing \textbf{160}, 85--92 (2015)

\bibitem{Wang20151}
Wang, J.J.Y., Wang, Y., Zhao, S., Gao, X.: Maximum mutual information
  regularized classification.
\newblock Engineering Applications of Artificial Intelligence \textbf{37}, 1--8
  (2015)

\bibitem{Wang2012963}
Wang, J.Y., Almasri, I., Gao, X.: Adaptive graph regularized nonnegative matrix
  factorization via feature selection.
\newblock In: Proceedings - International Conference on Pattern Recognition,
  pp. 963--966 (2012)

\bibitem{Wang2014258}
Wang, J.Y., Almasri, I., Shi, Y., Gao, X.: Semi-supervised transductive hot
  spot predictor working on multiple assumptions.
\newblock Current Bioinformatics \textbf{9}(3), 258--267 (2014)

\bibitem{wright2009robust}
Wright, J., Yang, A.Y., Ganesh, A., Sastry, S.S., Ma, Y.: Robust face
  recognition via sparse representation.
\newblock Pattern Analysis and Machine Intelligence, IEEE Transactions on
  \textbf{31}(2), 210--227 (2009)

\bibitem{Xu2015307}
Xu, Y., Fang, X., You, J., Chen, Y., Liu, H.: Noise-free representation based
  classification and face recognition experiments.
\newblock Neurocomputing \textbf{147}(1), 307--314 (2015)

\bibitem{Xu20121205}
Xu, Y., Shen, F., Zhao, J.: An incremental learning vector quantization
  algorithm for pattern classification.
\newblock Neural Computing and Applications \textbf{21}(6), 1205--1215 (2012)

\bibitem{Zhao2015677}
Zhao, S., Hu, Z.P.: A modular weighted sparse representation based on fisher
  discriminant and sparse residual for face recognition with occlusion.
\newblock Information Processing Letters \textbf{115}(9), 677--683 (2015)

\end{thebibliography}

\end{document}